\documentclass[letterpaper]{article} 
\usepackage{aaai2026}  
\usepackage{times}  
\usepackage{helvet}  
\usepackage{courier}  
\usepackage[hyphens]{url}  
\usepackage{graphicx} 
\usepackage{natbib}  
\usepackage{caption} 
\usepackage{algorithm}
\usepackage{algorithmic}
\usepackage{graphicx}
\usepackage{comment}
\usepackage{xcolor}

\usepackage{array}
\usepackage{booktabs}
\usepackage[most]{tcolorbox}
\usepackage{tabularx}
\usepackage{multirow}
\usepackage{enumitem}

\usepackage{pdfpages}

\usepackage{newfloat}
\usepackage{listings}
\DeclareCaptionStyle{ruled}{labelfont=normalfont,labelsep=colon,strut=off} 
\floatstyle{ruled}
\newfloat{listing}{tb}{lst}{}
\floatname{listing}{Listing}
\title{CyPortQA: Benchmarking Multimodal Large Language Models for Cyclone Preparedness in Port Operation}

\author{
    Chenchen Kuai\textsuperscript{\rm 1},
    Chenhao Wu\textsuperscript{\rm 2},
    Yang Zhou\textsuperscript{\rm 1},
    Bruce Wang\textsuperscript{\rm 1},
    Tianbao Yang\textsuperscript{\rm 1},\\
    Zhengzhong Tu\textsuperscript{\rm 1},
    Zihao Li\textsuperscript{\rm 1}\thanks{Corresponding authors},
    Yunlong Zhang\textsuperscript{\rm 1}\footnotemark[1]
}

\affiliations{
    \textsuperscript{\rm 1}Texas A\&M University, College Station, USA\\
    \textsuperscript{\rm 2}University of California, Los Angeles, USA\\
    \{mobility, yangzhou295, tianbao-yang, tzz, scottlzh\}@tamu.edu,\\
    \{chenhaowu\}@ucla.edu, \{bwang, yzhang\}@civil.tamu.edu
}

\usepackage{bibentry}

\begin{document}

\maketitle

\begin{abstract}

As tropical cyclones intensify and track forecasts become increasingly uncertain, U.S. ports face heightened supply-chain risk under extreme weather conditions. Port operators need to rapidly synthesize diverse multimodal forecast products, such as probabilistic wind maps, track cones, and official advisories, into clear, actionable guidance as cyclones approach. Multimodal large language models (MLLMs) offer a powerful means to integrate these heterogeneous data sources alongside broader contextual knowledge, yet their accuracy and reliability in the specific context of port cyclone preparedness have not been rigorously evaluated. To fill this gap, we introduce CyPortQA, the first multimodal benchmark tailored to port operations under cyclone threat. CyPortQA assembles 2,917 real-world disruption scenarios from 2015 through 2023, spanning 145 U.S. principal ports and 90 named storms. Each scenario fuses multi-source data (i.e., tropical cyclone products, port operational impact records, and port condition bulletins) and is expanded through an automated pipeline into 117,178 structured question–answer pairs. Using this benchmark, we conduct extensive experiments on diverse MLLMs, including both open-source and proprietary model. MLLMs demonstrate great potential in situation understanding but still face considerable challenges in reasoning tasks, including potential impact estimation and decision reasoning.

\end{abstract}

\begin{links}
    \link{Code\&Dataset}{https://github.com/ChenchenMobility/MLLM-Bench-CyPortQA}
\end{links}

\section{Introduction}

\begin{figure*}[!ht]
\centering
\includegraphics[width=0.99\textwidth]{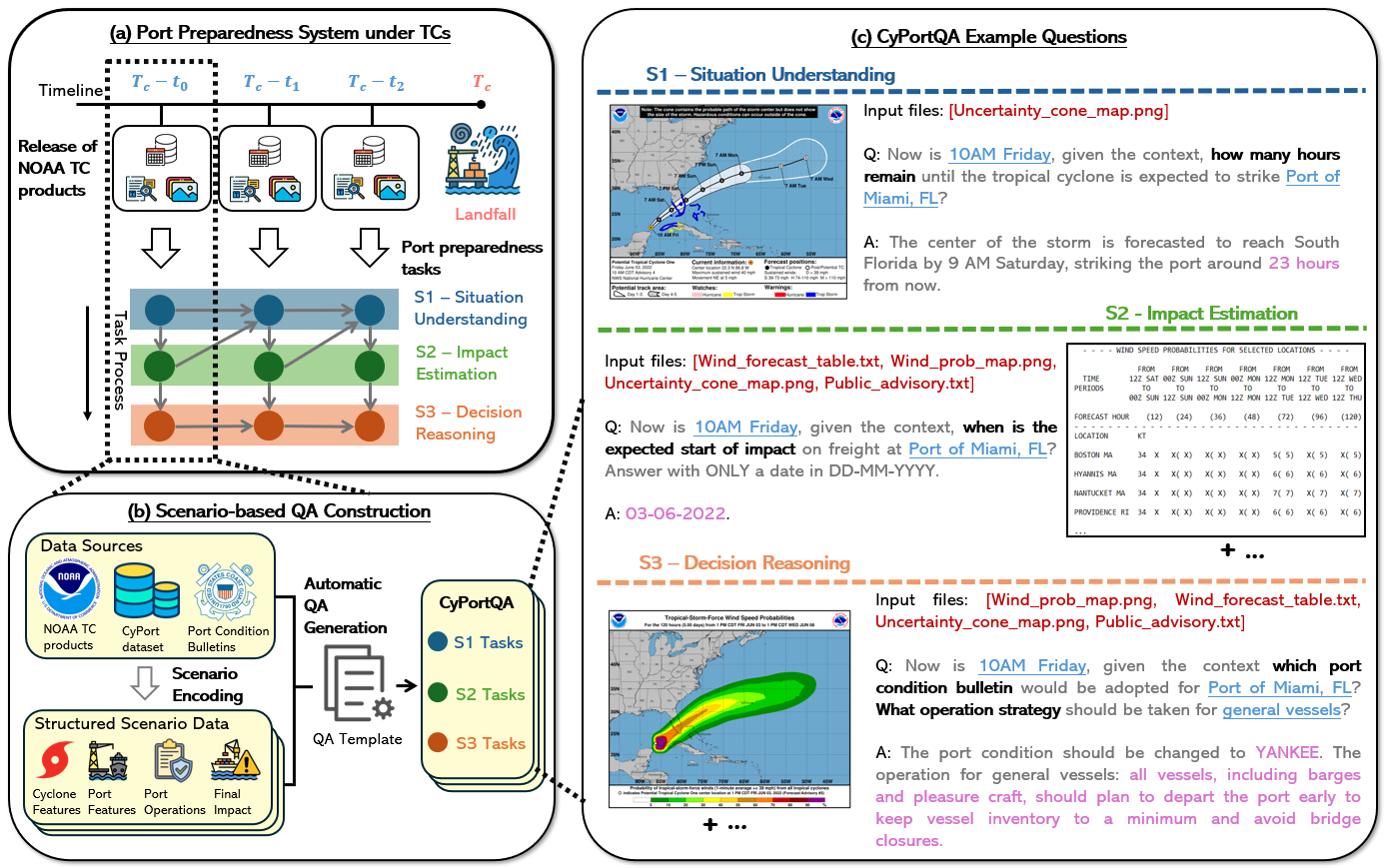} 
\caption{Port Preparedness Framework under Tropical Cyclones and the CyPortQA Benchmark. (a) The time-evolving port preparedness in response to TC, highlighting dynamic decision and key preparedness tasks. (b) Scenario-based QA construction pipeline in CyPortQA, sourcing NOAA TC products, operational performance data and port condition bulletins. (c) Representative CyPortQA examples across three preparedness tasks: S1 – Situation Understanding, S2 – Impact Estimation, and S3 – Decision Reasoning, each with corresponding multimodal inputs and question formats.}
\label{fig1}
\end{figure*}

Tropical cyclones\footnote{To be consistent, we use the term \textit{tropical cyclones} as the official designation encompassing hurricanes, typhoons, and cyclones.} (TCs) are becoming more intense and increasingly threaten U.S. ports, causing closures, infrastructure damage, and inland transport disruptions that lead to major economic losses \citep{Li2023PortRecovery,verschuur2023multi}. Over the past few decades, the number of cyclones rated Category 4 and Category 5 \footnote{According to the Saffir–Simpson Hurricane Wind Scale, which classifies cyclones based on sustained wind speeds.} has increased, and climate model forecast even stronger storms by the end of the century \citep{lipari2024amplified}. Meanwhile, increasing uncertainty on cyclone paths makes it more difficult for port operators to adjust operational plans dynamically and implement timely preparedness to reduce potential infrastructure damage and\ economic loss \citep{feehan2024ecosystem}.

Existing port-preparedness tools rely on diverse meteorological data sources to estimate storm impacts on operations and determine port condition bulletins \footnote{
 The USCG issues port condition bulletins such as Whiskey, X-Ray, Yankee, and Zulu under 33 CFR § 165.781~\citep{CFR_165_781}.}, which in turn regulate vessel movements and facility preparations~\citep{USCG2025SevereWeatherPlan}. Once a tropical cyclone reaches wind-force thresholds at sea, the National Hurricane Center (NHC) begins issuing forecast products at regular intervals, including probabilistic wind-speed maps, twelve-hour wind-forecast tables, narrative advisories on expected impacts, and graphical cones of uncertainty depicting possible storm tracks~\citep{NHC_Archive_2025,NHC_WindProb_Graphics}. Based on these products, NOAA (National Oceanic and Atmospheric Administration) and the USCG (U.S.\ Coast Guard) evaluate potential impacts on port operations. The USCG then applies a rule-based framework in which alerts are triggered when forecast wind speeds exceed predefined thresholds to determine preliminary operational decisions. Final port-condition bulletins are issued in consultation with port authorities. While this protocol provides a structured and actionable approach, it heavily depends on expert judgment. Additionally, other stakeholders, including vessel owners, terminal operators, and cargo interests, typically remain reactive and await official USCG notices before adjusting their operations ~\citep{Li2023PortRecovery,Balakrishnan2022EconRisk}.

Several challenges remain in model-driven real-time decision support for port preparedness, including improving interpretation of multifaceted forecasts, enhancing estimation of potential impacts, and ensuring reliability in decision-making~\citep{USCG2025SevereWeatherPlan}. Addressing these challenges calls for methods that can understand and reason over diverse and evolving data. Addressing these challenges requires models with robust understanding and reasoning abilities to process diverse, evolving data. Recent advances in MLLMs have demonstrated unprecedented capabilities in integrating heterogeneous inputs, such as images~\citep{jiang2025corvid}, structured tables~\citep{sui2024table}, and world knowledge~\cite{yu2023kola}, into coherent contextual outputs. By reasoning jointly across multiple modalities, MLLMs can capture complex relationships, adapt to incomplete or noisy information, and generate actionable insights, making them particularly promising for high-stakes applications in disaster management. To date, no study has systematically integrated dynamic tropical cyclone forecasts with ground-truth port operational impacts and official USCG port-condition bulletins. Without this integration, it is impossible to validate MLLM outputs against real-world outcomes, preventing in-depth performance assessment and slowing the development of automated decision-support tools. Hence, this study aims to address these gaps through the following contributions:


\begin{itemize}

\item Developing structured representations of real-world decision scenarios for port preparedness by integrating multisource data (e.g., meteorological forecasts, port operational impact, and USCG operational bulletins).

\item Building on these scenarios, we introduce \textit{CyPortQA}, the first multimodal benchmark dataset for port preparedness under cyclones, offering tasks in situational understanding, impact estimation, and decision reasoning.

\item Using \textit{CyPortQA}, we benchmark diverse MLLMs including both open-source and proprietary, assessing their performance and uncovering understanding and reasoning gaps in realistic cyclone-related port preparedness.

\end{itemize}

\section{Related Work}

\subsection{Models for Cyclone Preparedness}
Models used to estimate physical damage and disruption to ports caused by tropical cyclones are critical for both pre-event planning and post-event response within the disaster cycle~\citep{cai2025identifying}. The NOAA facilitates cyclone preparedness by publishing a suite of forecast models that provide timely and scientifically grounded predictions of storm tracks and intensities~\citep{noaacone,noaatext,demaria2013improvements,noaawindprob,boussioux2022hurricane}. It is equally critical to leverage these public forecast models to inform reliable cyclone impact estimation and support rapid decision making. However, most existing studies focus on long-term planning rather than real-time operational support. For example,  \citet{dhanak2021resilience} developed a simulation-based resilience assessment tool for strategic disaster planning. \citet{wang2025global} integrated causal inferences into autoregressive predictions for cyclone intensity forecasting. While these approaches offer important insights, they are not designed for dynamic decision support during active cyclone events. In contrast, \citet{Li2023PortRecovery} introduced a recommendation algorithm that models cyclones as users and ports as items, using historical interactions to predict the most likely impacted ports. However, this method relies on static hazard indicators such as maximum wind speed and minimum distance to the port, rather than real-time meteorological data, limiting its usefulness for dynamic operational decision making. In many cases, port authorities and terminal operators have to rely on manual interpretation of multimodal weather forecast products rather than integrating them into automated decision support systems~\citep{verschuur2020port}.

\subsection{MLLM in Natural Hazards}

Recent advances in MLLMs have demonstrated significant potential for natural hazard assessment by integrating heterogeneous data sources, such as textual reports and imagery, into a unified reasoning framework \citep{agarwal2020crisis}.  Existing work has been increasingly applied MLLMs across three interrelated pillars in natural hazard response: situation understanding \citep{hughes2025seeing,sun2023unleashing}, impact assessment \citep{sun2023unleashing,li2025llms,zhou2021analyzing}, and decision making \citep{chen2024integration,yin2024crisissense}. Previous studies have focused on leveraging the multimodal reasoning and situational awareness capabilities of MLLMs to address the immediate needs of hazard detection and emergency response under the impact of earthquakes \citep{ma2025multimodal}, hurricanes \citep{li2024mm,zhu2024flood}, and wildfires \citep{ramesh2025assessing,chen2025perceptions}. In addition, recent efforts have expanded the scope from a single hazard to multi-hazard environments~\citep{zhou2024assessing,wang2025disasterm3} that support a comprehensive understanding of cascading effects.  Despite these capabilities, the use of MLLMs in supporting port preparedness under the threat of tropical cyclones remains largely unexplored.

\section{Methodology}
This section presents the construction process of \textit{CyPortQA}. As shown in Figure~\ref{fig1}(a), the port-preparedness workflow reacts to real-time multimodal meteorological releases to assess the evolving situation and reason about potential impacts and regulation strategies. Based on this, We sample scenarios at key decision time points, pairing each with its corresponding weather data and operational observations. All inputs are encoded into a structured JSON formate, after which QA templates automatically generate question–answer pairs. This pipeline produces a consistent, scalable benchmark for evaluating multimodal understanding and reasoning in time-critical, high-stake port preparedness contexts.

\subsection{Data Sources}

\subsubsection{NOAA Tropical Cyclone Products}

\begin{figure}[tb]
\centering
\includegraphics[width=1.00\columnwidth]{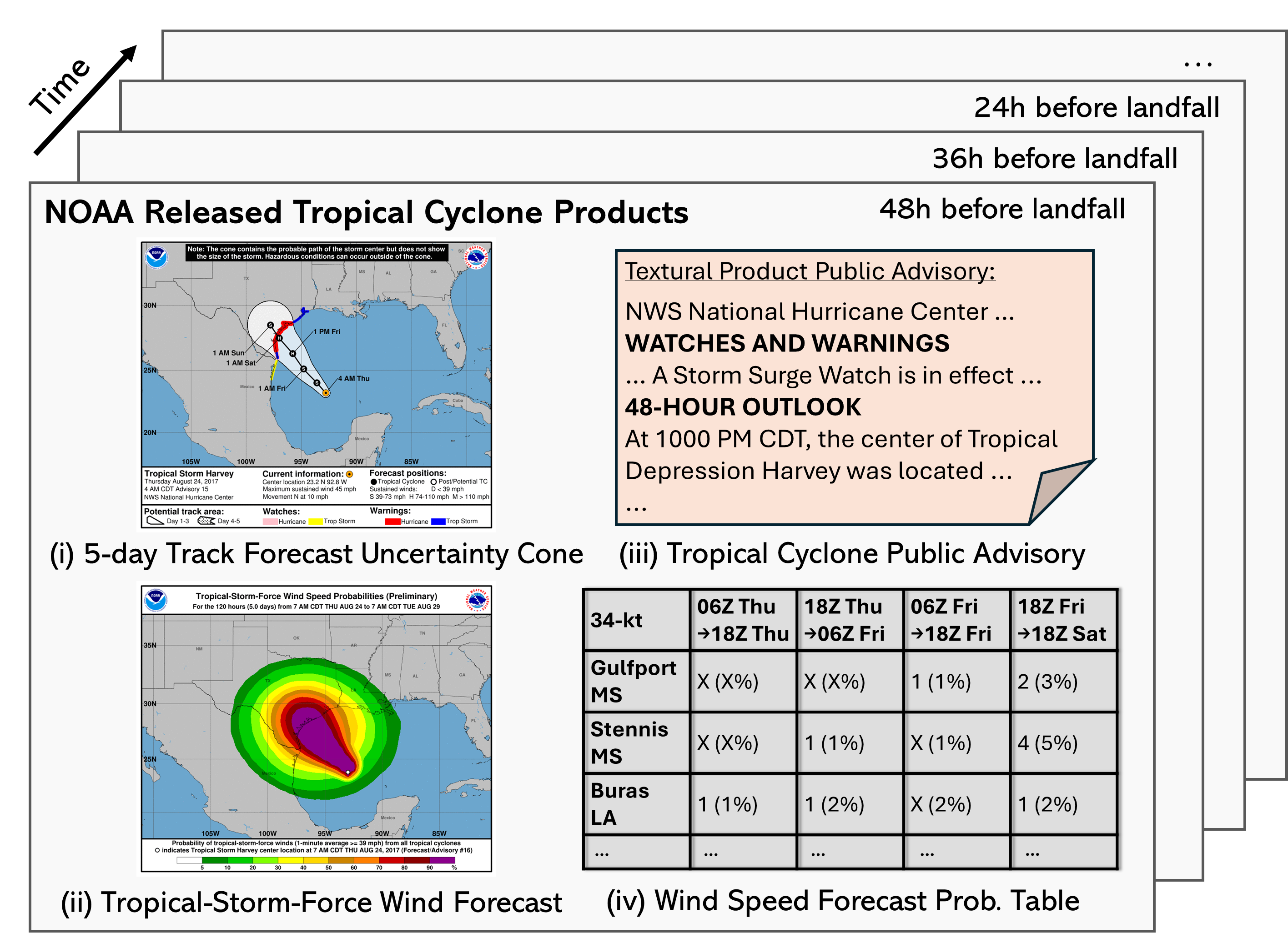}
\caption{Demonstration of NOAA released tropical cyclone weather products, example data from 2017 Harvey. The data is organized every 12 hours for port operation analysis.}
\label{fig:NOAAdata}
\end{figure}

NOAA tropical cyclone products provide updates every three hours as tropical cyclones move, with time-evolving cyclone information as well as meteorological conditions. As demonstrated in Figure \ref{fig:NOAAdata}, the data is organized in a time sequence following USCG evaluation time and is updated every 12 hours. At each data update interval, we extract four types of forecast products: 1) \emph{5-day Track-Forecast Cone with watch/warning region}, a map showing the projected storm path and coastal alert zones;  2) \emph{Tropical-Storm-Force Wind-Speed Probabilities Map}, showing the spatial likelihood of damaging winds; 3) Plain-text \emph{Tropical Cyclone Public Advisory}, summarizing position, intensity, motion, cyclone watches and warnings; and 4) \emph{Wind Speed Forecast Probabilities Table}, a fixed-width table listing the forecast wind-speed probabilities at selected locations and time bins.

Together, these maps, tables, and advisories represent the core information sources that port operators rely on for real-time situational awareness and decision-making.

\subsubsection{Port Operational Performance}

CyPort collects 2,197 tropical‑cyclone exposures affecting U.S. principal ports from 2015 to 2023~\citep{kuai2025us}. The port operational performance is represented by the commercial‑vessel AIS (Automatic Identification System) data captured by onboard navigation‑safety devices. For each port’s exposure to a tropical cyclone, CyPort documents commercial‑vessel activity and impact metrics, including the total reduction in vessel throughput, the onset date of disruption, and the recovery date when normal operations resume. This record offers a fine-grained view of how ports react to tropical cyclones, providing ground-truth data for potential impact estimation tasks.

\subsubsection{Port Condition Bulletins}
As tropical cyclone approaches, U.S. Coast Guard (USCG) districts issue different levels of \emph{Port‑Condition} bulletins, \emph{WHISKEY}, \emph{X‑RAY}, \emph{YANKEE}, or \emph{ZULU} that prescribe vessel and facility operations. We crawled these bulletins from official USCG X (formerly Twitter) channels (e.g., \texttt{@USCG}, \texttt{@USCGNortheast}, \texttt{@USCGSoutheast}), parsing for each alert the affected port(s), assigned port condition, release time, and the attached files detailing operational regulations. This information captures the regulatory actions faced by ports during tropical cyclones, serving as a reference for decision reasoning tasks. These data record the regulatory actions taken at time as tropical cyclone approaches.

Together, these three data sources describe real-time port operation scenarios as tropical cyclone approaching by combining real-time forecasts, observed port performance, and regulatory actions. They supply real-world test cases, complete with input meteorological data and verified outputs from ports to rigorously evaluate MLLMs.

\subsection{Data curation and Scenario Encoding}

Tropical cyclone information and port operational performances are sampled every twelve hours, starting 108 hours before the NHC’s declared landfall time and ending at landfall (\(T_{c}\)). Each sampling step collects NOAA tropical cyclone products and port condition bulletins, as well as port operational performance (i.e., CyPort dataset). CyPort dataset collects 1,927 cyclone–port exposures from 2015 to 2023. The port impact data is highly imbalanced: only about 21\% of these exposures lead to port disruption where freight throughput falls below the historical tenth percentile, and 6\% are accompanied by USCG Condition bulletins. To balance the classes, we retain all 410 exposures with impact on ports and select an equal number of exposures from ports within the defined impact buffer that exhibit no significant disruption to balance the classes ~\cite{kuai2025us}.

To assess ports’ real-time preparedness responses, we expand each of the 820 exposures across every 12-hour sampling point with available multimodal weather data, yielding 2,917 scenarios. Each scenario captures a specific cyclone and port at a certain time before landfall, along with weather forecasts, port condition bulletins, and port operational impact.

We convert each scenario's data, including the meteorological data, API-crawled port bulletins, and CyPort data records, into a unified JSON format using ChatGPT o3. Domain experts then independently cross-checked a random 10\% sample of scenarios and found strong alignment with the automated outputs, verifying the accuracy and reliability of the structured representations. Port performance impacts are taken directly from the structured CyPort dataset. Each JSON record specifies a cyclone identifier, a port identifier, and the number of hours before landfall, forming the metadata for our QA pipeline, Appendix 1 lists all scenario encoding prompts and JOSN template.

\subsection{Automatic QA Generation}

\begin{table*}[htbp]
  \centering
  \small
  \setlength{\tabcolsep}{4pt}
  \begin{tabular}{%
      >{\centering\arraybackslash}m{1.5cm}  
      >{\centering\arraybackslash}m{4cm}    
      >{\centering\arraybackslash}m{2cm}  
      >{\centering\arraybackslash}m{3cm}  
      >{\centering\arraybackslash}m{3cm}    
    }
    \toprule
    \textbf{Task}   & \textbf{Ability}           & \textbf{\# of QA}
      & \textbf{Input Modality}      & \textbf{Question Type} \\
    \midrule
    S1.1            & Spatial Awareness          & 46507
      & I / T / X            & TF / MC / SA              \\
    S1.2            & Temporal Understanding     & 40707
      & I / T / X            & TF / MC / SA             \\
    S1.3            & Exposure Interpretation    & 31885
      & I / T / X            & TF / MC / SA              \\
    S1.4            & Uncertainty Quantification & 29066
      & I / T / X            & TF / MC / SA              \\
    \midrule
    S2.1            & Time of Impact             & 5820
      & I + T + X                          & SA            \\
    S2.2            & Recovery Duration          &    2910   
      & I + T + X                            & SA                   \\
    S2.3            & Impact Severity          & 8730      
      & I + T + X                            & MC / SA                   \\
    \midrule
    S3.1            & Condition Alert        & 610  
      & I + T + X                          & MC               \\
    S3.2            & Operation Planning         & 822  
      & I + T + X                          & SA                  \\
    \bottomrule
  \end{tabular}
  
  \caption{Task breakdown in the CyPortQA benchmark. Each task targets a specific understanding or reasoning ability and is associated with a set of QA pairs grounded in multimodal inputs. Input modalities include I (Image), T (Table), and X (Text). Question types include TF (True/False), MC (Multiple Choice), and SA (Short Answer). The symbol `/` denotes alternatives (or), while `+` indicates combination (and).}
  \label{tab:portqa}
\end{table*}

\begin{table*}[t]
  \centering
  \small
  \setlength{\tabcolsep}{4pt}
  \begin{tabular}{@{}l
    *{4}{c}
    *{3}{c}
    *{3}{c}
    *{4}{c}
  @{}}
    \toprule
    & \multicolumn{4}{c}{\textbf{S1.1}}
    & \multicolumn{3}{c}{\textbf{S1.2}}
    & \multicolumn{3}{c}{\textbf{S1.3}}
    & \multicolumn{4}{c}{\textbf{S1.4}} \\
    \cmidrule(lr){2-5}
    \cmidrule(lr){6-8}
    \cmidrule(lr){9-11}
    \cmidrule(lr){12-15}
    \textbf{Model}
      & TF & MC  & SA‑Num & SA‑Desc
      & TF & MC  & SA‑Num
      & MC & SA‑Num & SA‑Desc
      & TF & MC  & SA‑Num & SA‑Desc \\
    \midrule
    LLaVA1.6     & 0.52  &  0.27  &  0.11  &       0.24 
                & 0.51  & 0.27   &   0.26      
                &  0.31  &    0.30     &        0.26
                &  0.49  & 0.26   &    0.30     &      0.01  \\
    LlaMA3.2       &  0.60 &  0.30  &   0.20    &    0.30    
                & 0.65  &  0.30  &   0.22      
                &   0.37 &   0.48      &      0.30  
                &  0.48  &  0.21  &     0.48    &    0.05    \\
    Gemma3       &  0.51 &   0.29 &  0.33       &       0.30 
                & 0.49  &  0.28  &   0.20      
                &  0.32  &   0.31      &       0.22 
                &   0.47 &  0.21  &    0.31       &   0.05   \\
    Qwen2.5‑VL   &  0.48 &  0.54  &   0.37      &     0.29   
                & 0.56  &  0.42  &    0.24     
                &  0.60  &   0.56      &       0.36 
                &  0.66  &  0.40  &   0.38      &    0.06    \\
    Mistral Small& 0.72  &  0.49  &    0.42   &       0.44 
                & 0.74  &  0.42  &   0.27      
                & 0.52   &  0.59       &     0.37   
                &  0.75  &  0.37  &     0.59    &   0.06     \\
    ChatGPT4o    & \textbf{0.81}  &  0.54  &     \textbf{0.53}    &     \textbf{0.54}
                &  \textbf{0.83} &  \textbf{0.54}  & \textbf{0.40}        
                & \textbf{0.61}  &    \textbf{0.63}    &  \textbf{0.45}      
                &  \textbf{0.83}  &   \textbf{0.55} &    \textbf{0.79}     &    \textbf{0.16}    \\
    Gemini2.5   &  0.74 &  \textbf{0.57}  &  0.40       & 0.47   
               & 0.78  & 0.51   &  0.29       
                &  0.60  &  0.59       &  0.38      
                &  0.82  &  0.41  &    0.59     &  0.15     \\
    \bottomrule
  \end{tabular}
  \caption{Performance comparison of MLLMs on situation understanding tasks across question-type. True/false and multiple-choice items are scored by exact accuracy. Short-answer numeric (SA-Num) items use tolerance-based accuracy. Text-description items (SA-Desc) are graded by an LLM judge on a 0–1 scale ($\uparrow$).}
  \label{tab:understanding-transposed}
\end{table*}

The automatic QA generation process is grounded in a generalized template framework. For each scenario, we employ a curated set of 48 QA templates, into which scenario metadata can be systematically inserted to generate precise QA pairs. Each template is designed to ensure that the answer can be directly validated from the scenario encoding, thereby preserving label accuracy while evaluating the MLLM’s ability to understand and reason over one or more input modalities. The resulting QA pairs collectively span a wide range of tasks aligned with key stages of preparedness: \textit{situation understanding (S1)}, \textit{impact estimation (S2)}, and \textit{decision reasoning (S3)}.

QA generation runs only when a template’s required fields are present in a scenario’s encoded JSON. Whenever the criteria are met, the system instantiates the QA pair and adds it to CyPortQA. Because port-condition bulletins are not issued in most scenario, S3-type decision reasoning questions receive sparser coverage. In Table \ref{tab:portqa}, CyPortQA ultimately contains 117,178 QA pairs spanning all stages of the cyclone-preparedness workflow. To mitigate potential template bias, each question was paraphrased into several semantically equivalent variants for consistency testing. QA templates and paraphrasing details are provided in Appendix 2, and the design criteria for each stage are presented below. 

\subsubsection{Situation Understanding (S1)}

Understanding-oriented QA templates are designed around four key dimensions essential to accurate disaster resilience insights: \textit{spatial awareness}, \textit{temporal understanding}, \textit{exposure interpretation}, and \textit{uncertainty quantification}. These dimensions align with core components of situational awareness during disaster preparedness and response~\cite{lei2025harnessing}.

Spatial awareness items test whether an MLLM can reason about geographic relationships from meteorological products, such as whether a port lies within a cyclone’s forecast cone or inside a specified wind-probability zone. Because preparedness is time-critical, the QA templates also probe temporal understanding (e.g., hours remaining until landfall). Exposure interpretation templates assess how well the model gauges a port’s level and seriousness of exposure. Finally, uncertainty quantification QAs evaluate the model’s ability to work with probabilistic information, such as estimating the likelihood that a port will be affected.

\begin{figure}[ht]
\centering
\includegraphics[width=0.88\columnwidth]{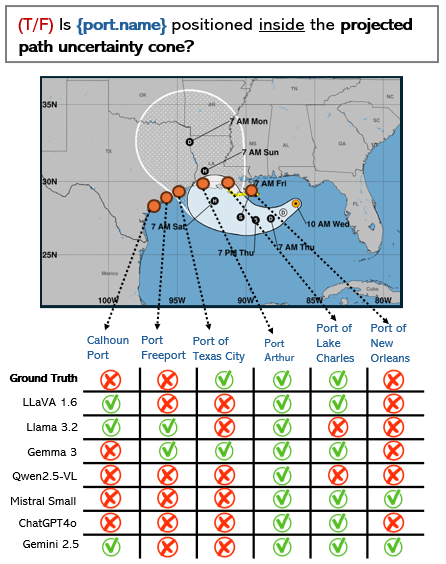}
\caption{
Spatial Awareness \& Exposure Interpretation Gaps. A tick indicates a 'yes' response (the port lies within the cyclone’s uncertainty cone), while a cross indicates a 'no' response (the port lies outside).
}
\label{fig:Fig3}
\end{figure}

\subsubsection{Impact Estimation (S2)}

Impact estimation QA templates are categorized into three distinct subtasks, \textit{time of impact}, \textit{recovery duration}, and \textit{impact severity}, that examine an MLLM to fuse evolving weather data with port attributes and forecast when freight operations will be disrupted, how long recovery will take, and how intense the disruption will be. These ground-truth labels are derived from vessel AIS records within the CyPort dataset, which capture disruptions in freight activity during tropical cyclones. Potential impact estimation in the benchmark verifies whether MLLMs can translate evolving weather into freight impact predictions, which helps commercial vessels reroute proactively and minimize economic loss~\cite{Li2023PortRecovery}.

\subsubsection{Decision Reasoning (S3)}
Decision reasoning is the final task of the benchmark, encompassing two critical tasks: determining the appropriate port condition bulletin to issue, and generating port-level operational strategies. These strategies include restrictions on general vessel traffic, ocean-going commercial vessels, and port facility operations as the cyclone approaches. 
Since ports serve as critical links in maritime supply chain, post-cyclone decision making requires precise calibration, as both over-reaction and under-reaction can lead to cascading disasters. Therefore, two additional metrics, over-reaction and under-reaction rates, are incorporated in this stage of evaluation.

\begin{table*}[t]
  \centering
  \small
  \setlength{\tabcolsep}{10pt}
  \begin{tabular}{@{}l
      *{4}{c}  
      *{2}{c}  
    @{}}
    \toprule
    & \multicolumn{4}{c}{\textbf{S2}} 
      & \multicolumn{2}{c}{\textbf{S3}} \\
    \cmidrule(lr){2-5}\cmidrule(lr){6-7}
    \textbf{Model}
      & S2.1
      & S2.2
      & S2.3 (MC)
      & S2.3 (SA‑Num)
      & S3.1
      & S3.2 \\
    \midrule
    LLaVA1.6             &  0.11 & 0.09  & 0.26  & 0.16  & 
    0.22 (0.43/0.27)&  0.36 (0.63/0.11) \\
    LlaMA3.2               &  0.30 & 0.27  & 0.26  & 0.24  & 
    0.29 (0.61/0.05) &  0.41 (0.43/0.17) \\
    Gemma3               &  0.41 & 0.28  &  0.29 &  0.31 &             0.10 (0.37/0.10)        &  0.27 (0.68/0.12)\\
    Qwen2.5‑VL           &  0.17 &  0.43 & 0.46  & 0.44  &                0.21 (0.62/0.17)     &   0.40 (0.65/0.11)\\
    Mistral Small        & 0.12  &  0.12 & 0.24  & 0.28  &           0.12 (0.09/0.07)      &   0.05 (0.70/0.10)\\
    ChatGPT4o            &  0.55 & 0.70  & 0.44  &  0.34 &           \textbf{0.35} (0.37/0.27)          & \textbf{0.53}  (0.62/0.02)  \\
    Gemini2.5          & \textbf{0.69}  & \textbf{0.72}  &  \textbf{0.48} & \textbf{0.44}  & 0.22 (0.58/0.22)      &  0.41 (0.67/0.08) \\
    \bottomrule
  \end{tabular}
  \vspace{0.5em}
  \caption{Performance comparison of MLLMs on reasoning tasks across question-type. Columns S3.1–S3.2 as decision reasoning questions results are demonstrated as
\emph{Acc} (\emph{UR}/\emph{OR}), where \emph{Acc} is mean accuracy or score,
\emph{UR} is the \underline{under-reaction} rate (MLLM recommends a less stringent regulation than required),
and \emph{OR} is the \underline{over-reaction} rate (MLLM recommends a more stringent regulation than required).}
  \label{tab:reasoning-comparison}
\end{table*}

\section{Experiments}

We conduct experiments on the proposed CyPortQA benchmark to evaluate MLLM performance. Diverse baselines (five open-source and two proprietary models) are tested to ensure comprehensive model coverage.

\begin{itemize}
  \setlength\itemsep{0pt}        
  \item \texttt{LLaVA-1.6-7B}~\cite{liu2023improvedllava}
  \item \texttt{Llama-3.2-Vision-11B}~\cite{dubey2024llama}
  \item \texttt{Gemma-3-12B}~\cite{team2025gemma}
  \item \texttt{Qwen-2.5-VL-7B}~\cite{bai2025qwen2}
  \item \texttt{Mistral-Small-3.1-24B}~\cite{mistraldocu}
  \item \texttt{ChatGPT-4o}~\cite{openai2025}
  \item \texttt{Gemini-2.5-Flash-Lite}~\cite{comanici2025gemini}
\end{itemize}


\subsection{Questions and Evaluation Metrics}
\subsubsection{Close-ended questions}
This question type includes True-or-False and multiple-choice formats, each with a single correct answer. Accuracy is used as the evaluation metric, calculated as the proportion of MLLM responses that exactly match the ground-truth answer.

\subsubsection{Open-ended Questions}
Open-ended questions require short answers, which are either numeric values (i.e., specific numbers, dates, or durations) or descriptive responses (e.g., operational suggestions). To evaluate the numeric outputs, we apply tolerance-based accuracy \citep{tian2025nuscenes}, where a prediction is considered correct if it falls within a predefined acceptable range based on domain understanding (e.g., Recovery duration is considered correct when it falls within ± 1 day of the ground truth). For descriptive answers, we adopt Multi-LLMs-as-Judge approach, where four top-ranked LLM judges \citep{LiEtAl2024_ArenaHard_BenchBuilder} were used: GPT-o3, Gemini-2.5-Pro, Claude-Sonnet-4, and GPT-o4-mini. The final evaluation score of the Multi-LLM judge is computed as the average across all four models. For S3.2 task, the judges are also required to classify responses to one of the following three categories: under-reaction, proportionate reaction, or over-reaction. The final classification is determined by majority vote. Implementation details for LLM judges are provided in Appendix 3.

\subsection{Evaluation of situation understanding (S1)}


The results across all situation understanding dimensions are presented in Table~\ref{tab:understanding-transposed}. Top-performing MLLMs demonstrate high accuracy, with spatial awareness, temporal understanding, and uncertainty quantification all exceeding 80\% accuracy for True/False questions. This highlights the potential ability of MLLMs in understanding the Tropical Cyclone scenarios and extracting key port preparedness information. Descriptive uncertainty questions score poorly, revealing that the models struggle to convert multimodal inputs into precise probabilistic risk statements.

Overall, proprietary models outperform open-source counterparts. ChatGPT-4o achieves the highest accuracy across tasks involving all abilities. Gemini 2.5 Flash-Lite shows competitive performance on spatial understanding multiple-choice questions and performs comparably well. Among open-source MLLMs, Mistral-Small and Qwen-2.5-VL demonstrate stronger performance compared to LLaVA-1.6, LLaMA-3.2, and Gemma-3, particularly in exposure interpretation and uncertainty quantification tasks.


\subsubsection{Error Analysis for situation understanding}
Figure~\ref{fig:Fig3} shows that the MLLMs correctly identify Port Author, which is located well within the uncertainty cone, demonstrating their ability to perform coarse-grained spatial localization. However, performance degrades when ports are located on or near the edge of the uncertainty cone. For instance, only Gemma-3 correctly classifies Port of Texas City, underscoring how boundary ports confound other models. Although geographically distant from the cone, Calhoun Port are incorrectly labeled 'within' by three models. Smaller ports may have less geographical knowledge in MLLMs, and spatial-awareness questions involving them are more often misinterpreted. Taken together, these errors indicate that MLLMs still lack the fine-grained spatial understanding in port operations.

\begin{figure}[ht]
\centering
\includegraphics[width=0.99\columnwidth]{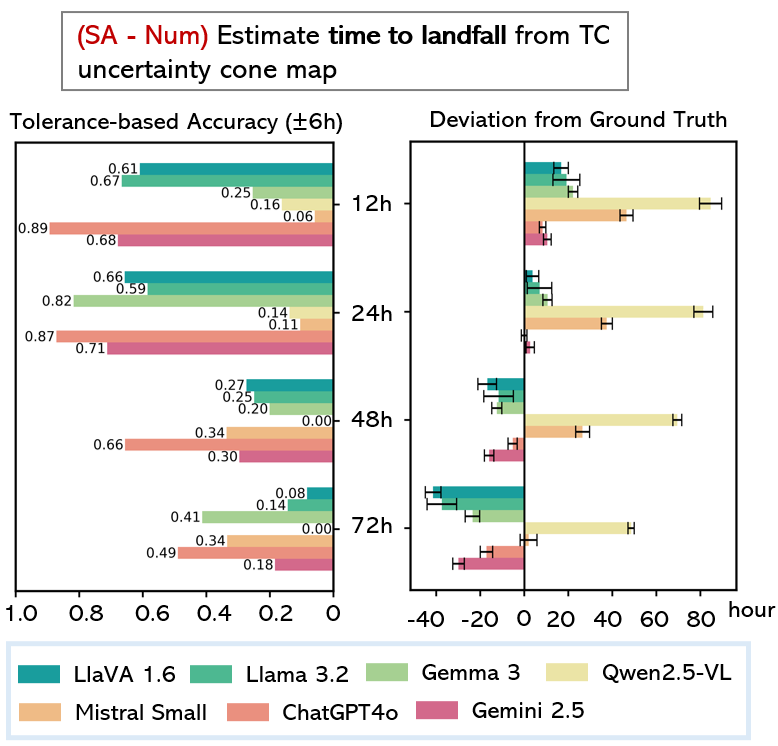}
\caption{Temporal-understanding gaps. Performance aggregated at 72, 48, 24, and 12 h before landfall. Left panel: tolerance-based accuracy; right panel: mean deviation from ground truth with 95 \% confidence intervals.}
\label{fig:Fig4}
\end{figure}

Temporal understanding (Figure \ref{fig:Fig4}) reveals another gap. Across the selected lead times (12h, 24h, 48h, and 72h to landfall), prediction accuracy tends to improve as the time approaches landfall due to the reduced forecast uncertainty in later stages. However, the smallest deviation from ground truth is reached at 24 h rather than 12h for most models (e.g., ChatGPT-4o, Gemini-2.5, and LLaVA-1.6). These MLLMs tend to predict a time close to a full day (i.e. 24h). This is revealed by earlier findings that MLLMs exhibit a prior distribution or bias which favors certain outputs or decisions \citep{McCoy2024Embers}. This imprecision or ``preference'' can lead to inaccurate judgments in port preparedness decisions.

\begin{figure}[ht]
\centering
\includegraphics[width=1.00\columnwidth]{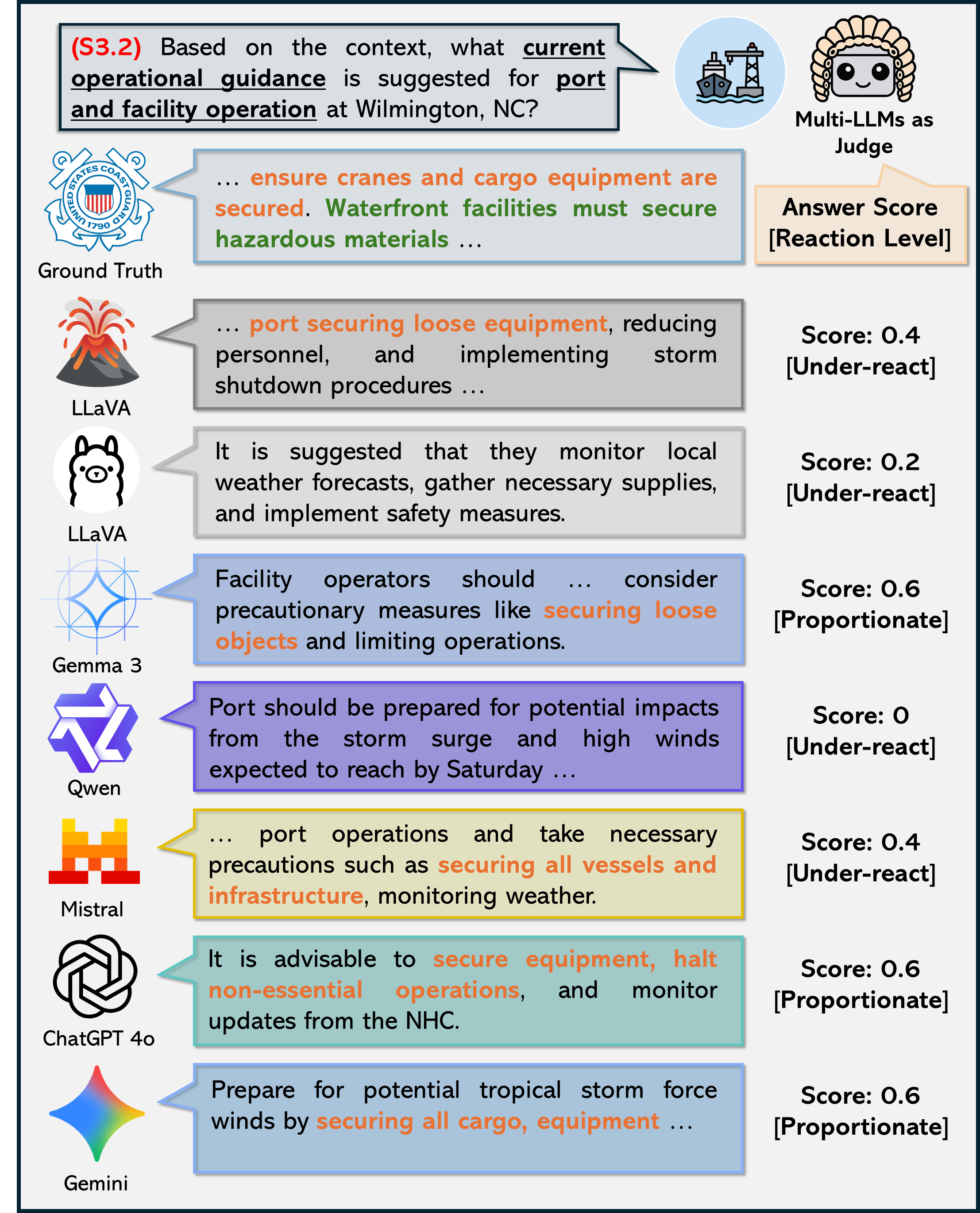}
\caption{Responses from MLLMs for decision reasoning tasks (port and facility operation instructions) under a single scenario. Evaluation results from the LLM-as-a-judge include a numerical score and classify each response as either an under-reaction, normal reaction, or over-reaction.}
\label{fig:Fig5}
\end{figure}

\subsection{Evaluation of reasoning tasks (S2\&S3)}

Results for reasoning tasks are shown in Table~\ref{tab:reasoning-comparison}. The evaluated MLLMs show clear differences in impact estimation tasks (S2), with Gemini outperforming all other models in multiple-choice accuracy. Furthermore, Qwen2.5-VL demonstrates better performance than ChatGPT-4o and all other open-source models. For S3 tasks, ChatGPT-4o shows a much higher likelihood of issuing the correct bulletin at the appropriate time, whereas its correctly issued bulletin still remains insufficient for real-world operational deployment. In addition, most models tend to underestimate the severity of situations, frequently assigning lower level bulletins when confronted with potential threats. This pattern is particularly evident in Gemini-2.5, LLaMA-3.2, and Qwen-2.5-VL. In practical applications, such under-reactions can lead to delayed or insufficient preparation, potentially exposing vessels, ports, and key infrastructure to unexpected damage.

\subsubsection{Error Analysis for decision reasoning}
Figure~\ref{fig:Fig5} presents a scenario that reveals reasoning gaps in MLLM-generated operational responses. In this case, none of the baseline models correctly address the regulatory requirements needed for coordinated port and facility preparedness. In particular, all models fail to mention ``securing waterfront facility'' that form a key part of port readiness protocols.

The missing portions of the responses reveal insufficient knowledge of waterfront facility regulations, core guidance for port operations. Domain knowledge thus becomes a key bottleneck preventing MLLMs from effectively identifying the specific procedures and interdependent actions. This deficiency further results in vague, high-level responses that generalize or oversimplify operational requirements, making the operations underestimating the urgency of real-world decisions. These results highlight the importance of grounding model outputs in accurate situational understanding and domain-specific knowledge. Effective decision support requires MLLMs that integrate domain expertise to translate weather observations into precise, actionable guidance.

\section{Conclusion}
We present CyPortQA, a multimodal benchmark for cyclone preparedness in port operations. Constructed from nine years of real-world disruption scenarios across 145 major U.S. ports and 90 named cyclones, it integrates verified data from NOAA tropical cyclone products, port operational impact records, and USCG port-condition bulletins. Using CyPortQA, we evaluate seven MLLMs (both open-source and proprietary) on three escalating tasks: situation understanding, impact estimation, and decision reasoning. Our experiments indicate that proprietary MLLMs outperform open-source models on both understanding and reasoning tasks. While these models demonstrate strong potential in situation awareness to support port preparedness, they reveal notable limitations in advanced reasoning tasks such as precise impact estimation and actionable decision support. We release CyPortQA to inspire further research on reliable, LLM-assisted emergency decision-support tools that enhance critical infrastructure resilience and operational effectiveness, especially under natural disasters.

\textbf{Limitations and Future Works:} Although CyPortQA provides a comprehensive benchmark, it is constructed from U.S.-based tropical cyclone cases. As a result, its applicability to regions with different cyclone patterns and port operations has yet to be tested. Future work should focus on expanding the dataset to include global ports and diverse hazard events, enabling cross-regional evaluation.

\section{Acknowledgments}
We thank Mr. Blakemore of the USCG for his invaluable insights on port disaster preparedness. We also thank undergraduate Brady Foster for data collection and curation. This research is funded by the Center for Freight Transportation for Efficient \& Resilient Supply Chain (FERSC) through the U.S. Department of Transportation (69A3552348338). All datasets used in this study (i.e., AIS, NOAA, CyPort, and USCG Port Condition bulletins) are publicly available and redistributable, with no privacy concerns.

\bibliography{aaai2026}
\clearpage
\onecolumn
\includepdf[pages=-]{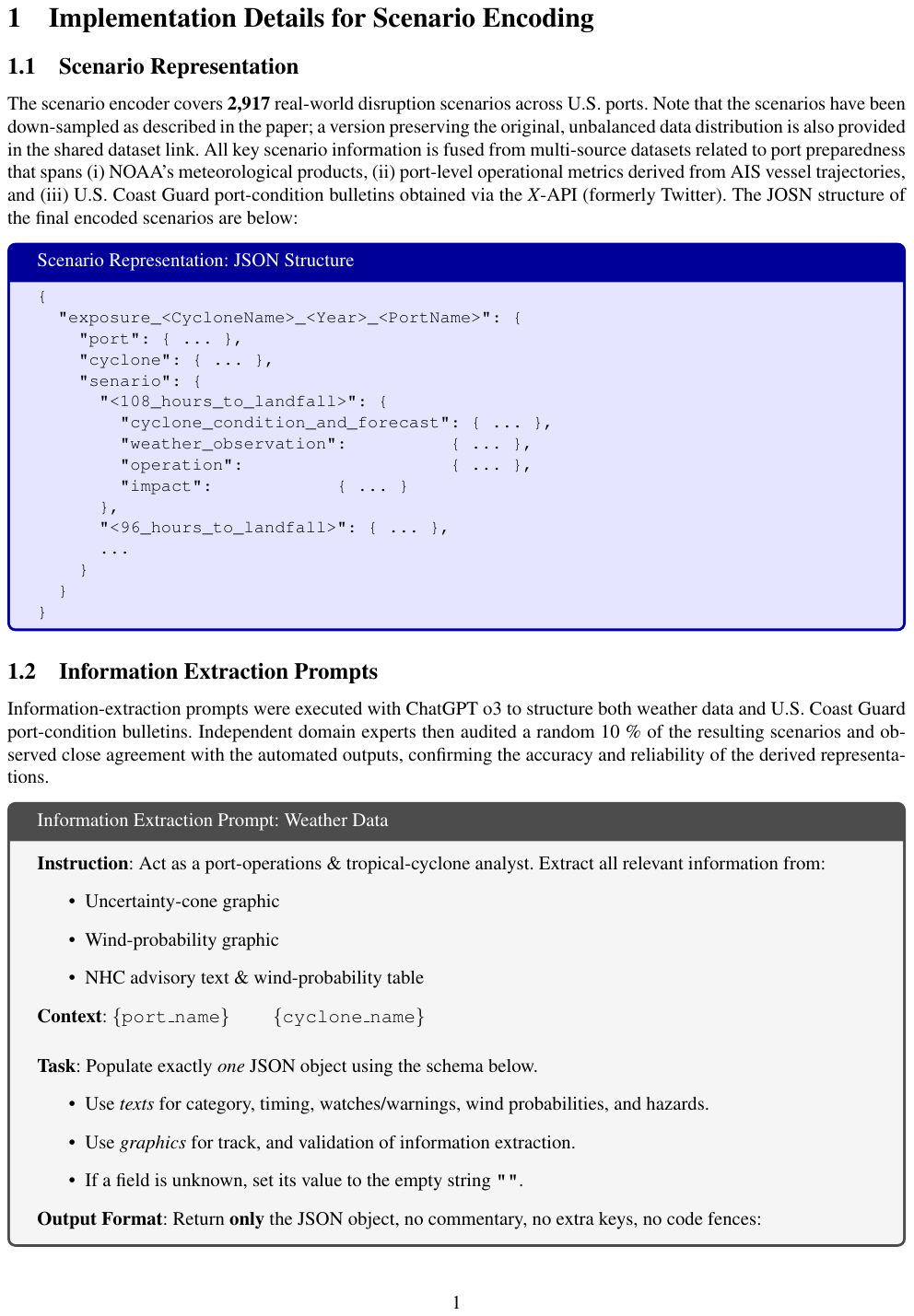}

\end{document}